\documentclass[letterpaper, 10 pt, conference]{ieeeconf}  % Comment this line out if you need a4paper

\IEEEoverridecommandlockouts                              % This command is only needed if 
\usepackage{graphics} % for pdf, bitmapped graphics files
\usepackage{graphicx}
\usepackage{xcolor}
\usepackage{subcaption}
\usepackage{makecell}

\title{\LARGE \bf
NeuFlow: Real-time, High-accuracy Optical Flow Estimation on Robots Using Edge Devices
}

\author{Zhiyong Zhang$^{1}$, Huaizu Jiang$^{2}$, Hanumant Singh$^{1}$ % <-this % stops a space
\thanks{$^1$Department of Electrical and Computer Engineering, Northeastern University, Boston, MA 02115. {\tt\small zhang.zhiyo@northeastern.edu}
{\tt\small ha.singh@northeastern.edu}
$^2$Khoury College of Computer Sciences, Northeastern University, Boston, MA 02115. {\tt\small h.jiang@northeastern.edu}}
}

\begin{document}

\maketitle
\thispagestyle{empty}
\pagestyle{empty}

%%%%%%%%%%%%%%%%%%%%%%%%%%%%%%%%%%%%%%%%%%%%%%%%%%%%%%%%%%%%%%%%%%%%%%%%%%%%%%%%
\begin{abstract}

Real-time high-accuracy optical flow estimation is a crucial component in various applications, including localization and mapping in robotics, object tracking, and activity recognition in computer vision. While recent learning-based optical flow methods have achieved high accuracy, they often come with heavy computation costs. In this paper, we propose a highly efficient optical flow architecture, called NeuFlow, that addresses both high accuracy and computational cost concerns. The architecture follows a global-to-local scheme.
Given the features of the input images extracted at different spatial resolutions, global matching is employed to estimate an initial optical flow on the 1/16 resolution, capturing large displacement, which is then refined on the 1/8 resolution with lightweight CNN layers for better accuracy. We evaluate our approach on Jetson Orin Nano and RTX 2080 to demonstrate efficiency improvements across different computing platforms. We achieve a notable 10×-80× speedup compared to several state-of-the-art methods, while maintaining comparable accuracy. Our approach achieves around 30 FPS on edge computing platforms, which represents a significant breakthrough in deploying complex computer vision tasks such as SLAM on small robots like drones. The full training and evaluation code is available at https://github.com/neufieldrobotics/NeuFlow.

\end{abstract}

%%%%%%%%%%%%%%%%%%%%%%%%%%%%%%%%%%%%%%%%%%%%%%%%%%%%%%%%%%%%%%%%%%%%%%%%%%%%%%%%
\section{Introduction}

Optical flow estimation, a fundamental task in computer vision \cite{fortun2015optical}, plays a pivotal role in various applications such as object tracking \cite{shin2005optical, kale2015moving}, motion analysis \cite{aggarwal1999human}, scene understanding \cite{saleemi2010scene}, and visual odometry \cite{teed2021droid, muller2017flowdometry}. Optical flow refers to the distribution of apparent velocities of movement of brightness patterns in an image \cite{horn1981determining}, which can result from the relative motion of objects and the viewer \cite{gibson1950perception}, \cite{gibson1966senses}. Real-time optical flow estimation, in particular, holds immense significance in scenarios requiring quick and accurate analysis of dynamic scenes \cite{liu2021rdmo}, ranging from robotics \cite{whelan2016elasticfusion} to autonomous driving systems \cite{behl2017bounding}, augmented reality \cite{jain2015overlay}, and beyond \cite{chao2014survey}.

\begin{figure}[t]
\begin{center}
\includegraphics[width=\linewidth]{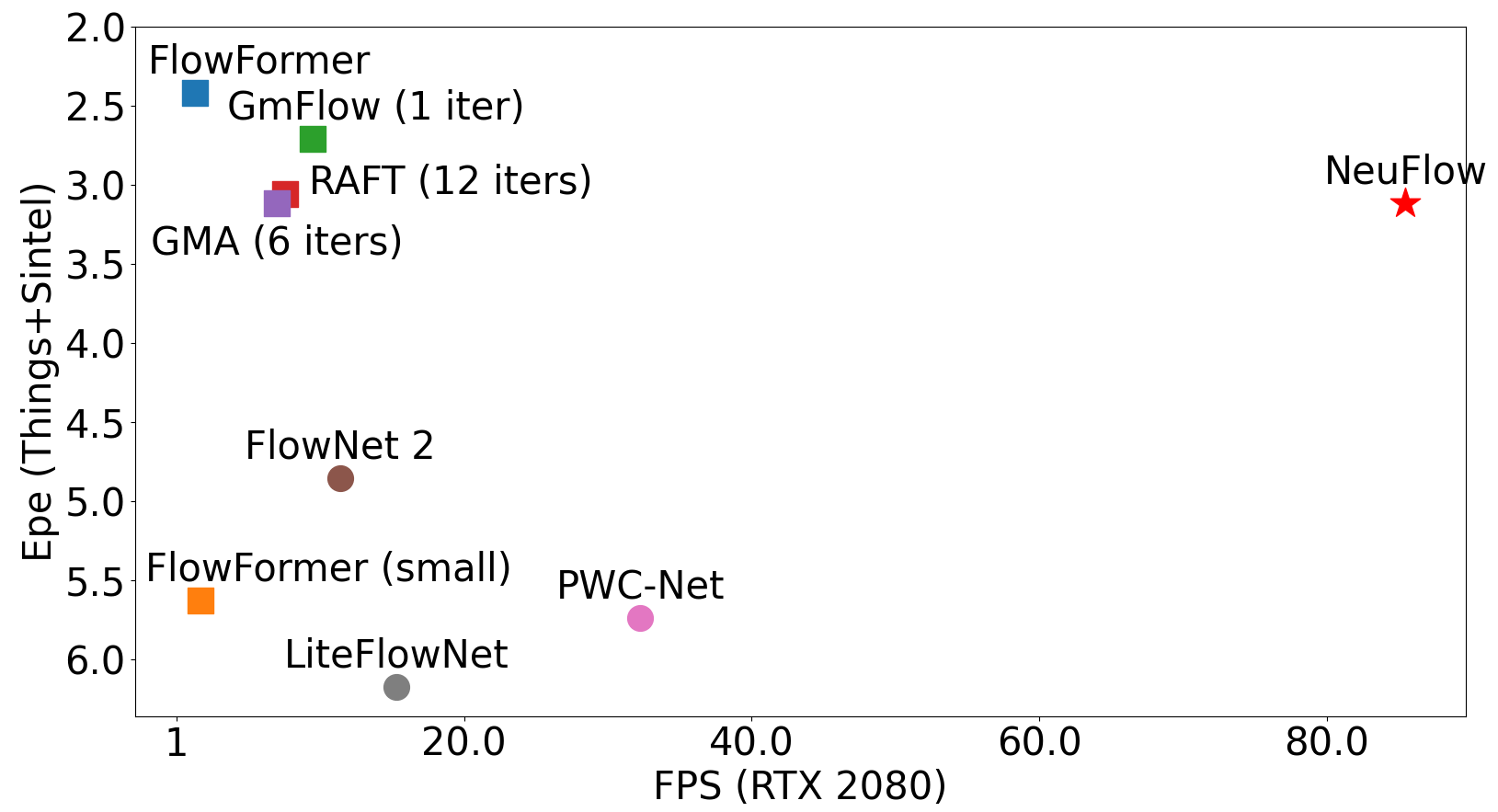}
\end{center}
\caption{End point error (EPE) v.s. frame per second (FPS) throughput on a common computing platform (Nvidia RTX 2080). Individual points represents a broad class of optical flow methods. Our algorithm is comparable in accuracy but significantly better (close to an order of magnitude) in terms of its computational complexity. All models trained solely on FlyingThings and FlyingChairs.}  
\label{epe_fps_1_1}
\end{figure}

\begin{figure}[h]
\begin{center}
\includegraphics[width=\linewidth]{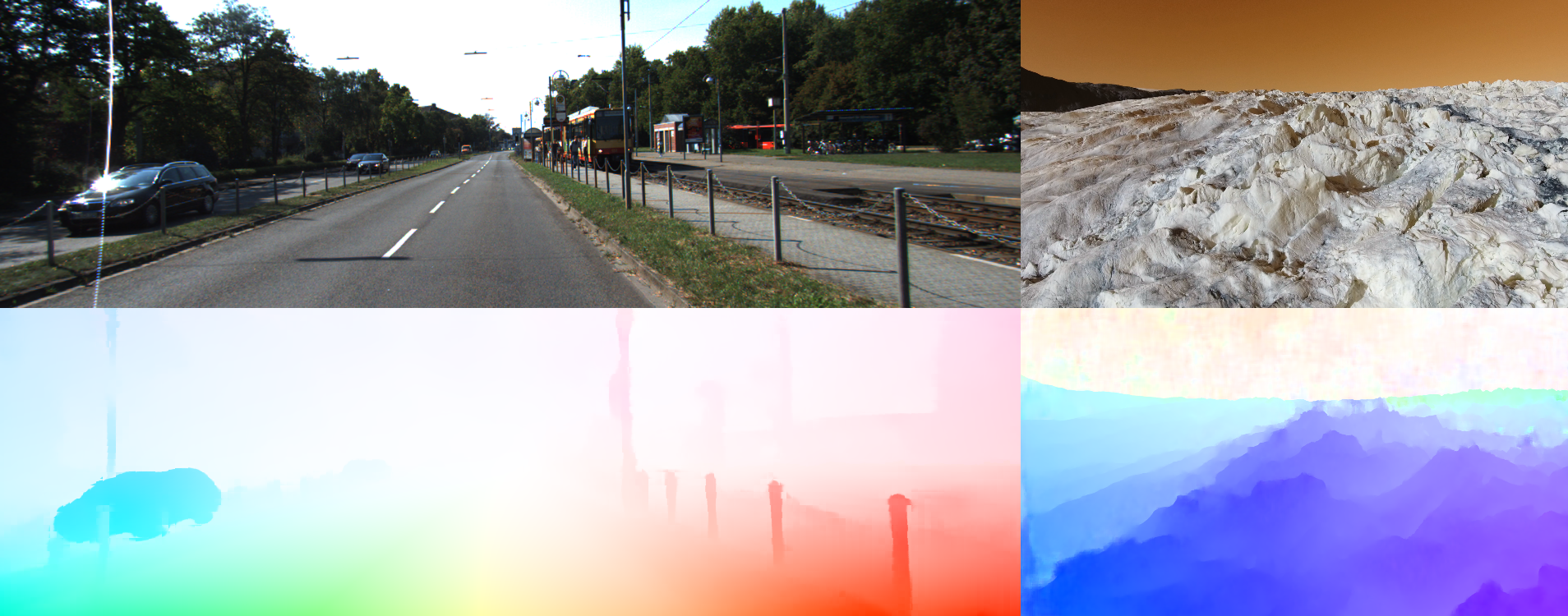}
\end{center}
\caption{Optical flow results of NeuFlow: on the left is a result from the standard KITTI dataset. On the right are results from a UAS flight overlow-contrast glacier images in the Arctic. Our approach is notable for both computational efficiency and speed as well as accuracy, as shown in Fig. \ref{epe_fps_1_1}.}
\label{optical_flow_vis}
\end{figure}

In recent years, significant advancements have been made in the development of algorithms and techniques aimed at achieving high-accuracy optical flow estimation \cite{dosovitskiy2015flownet, ilg2017flownet, teed2020raft, xu2022gmflow, huang2022flowformer}. Starting from FlowNet \cite{dosovitskiy2015flownet}, learning-based optical flow methods have emerged to learn features for matching instead of relying on hand-crafted features like Lucas-Kanade \cite{lucas1981iterative} or SIFT \cite{lowe2004distinctive, liu2008sift}. However, early optical flow methods still suffer from two major problems: large displacement and ambiguity \cite{xu2022gmflow}. Recent deep learning methods \cite{teed2020raft, jiang2021learning, xu2022gmflow, sui2022craft, huang2022flowformer, shi2023flowformer++} have made strides in addressing these issues to some extent, at the expense of computation time.

\begin{figure*}[ht]
\begin{center}
\includegraphics[width=\textwidth]{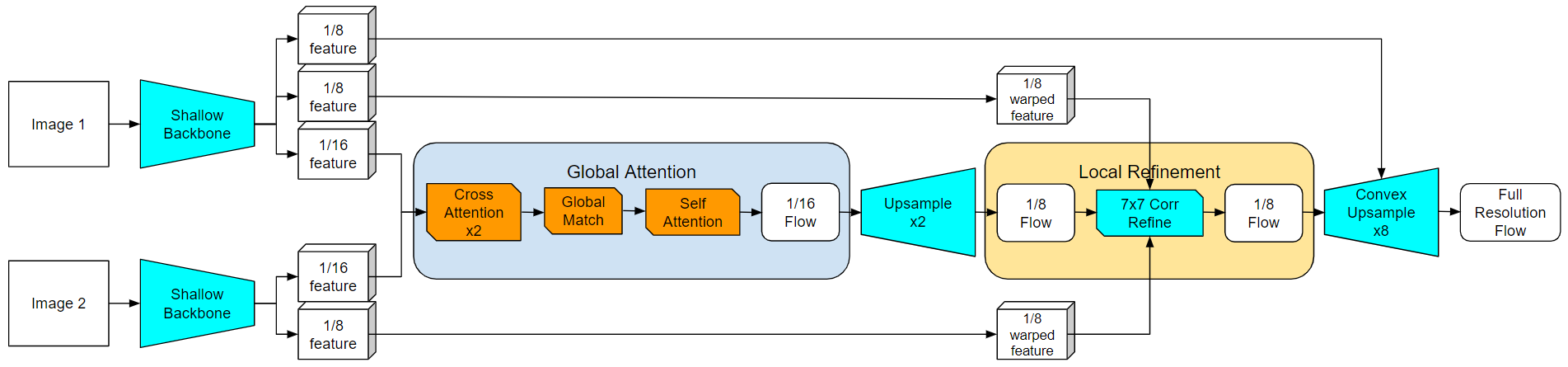}
\end{center}
\caption{NeuFlow Architecture: We begins with a shallow CNN backbone. The backbone outputs feature vectors at 1/8 and 1/16 scale for both images. The feature vectors at 1/16 scale are then fed into two cross-attention layers for global matching. The resulting flow is passed into a self-attention layer for flow propagation based on feature self-similarity. Subsequently, the flow is upsampled to obtain 1/8 resolution flow. We wrap the 1/8 features with the flow and perform local refinement within a 7x7 window. The refined 1/8 flow is then upsampled to obtain full-resolution flow using a convex upsampling module, which additionally requires 1/8 features from image one.}
\label{neuflow_arct}
\end{figure*}

Early optical flow methods rely on CNNs (Convolutional Neural Networks) and local correlations of image features, which can only capture small-displacement pixel movement due to the restricted range of operation of these techniques \cite{xu2022gmflow}. Recent solutions, such as RAFT \cite{teed2020raft} and GMA \cite{jiang2021learning}, use iterative methods to mitigate such problems. Transformer-based approaches, like GmFlow \cite{xu2022gmflow} or CRAFT \cite{sui2022craft}, leverage global attention \cite{vaswani2017attention} layers to address the issue. However, iterative methods necessitate numerous iterations to estimate large-displacement optical flow \cite{teed2020raft} and global attention computes correlations between each pair pixels across two images, both resulting in significant computational costs \cite{sui2022craft}.

Another challenge is ambiguity, which includes occlusions and textureless regions \cite{tu2019survey}, is typically addressed by aggregating pixels that likely belong to the same object \cite{jiang2021learning}. Early optical flow methods, constrained by the limited receptive field of CNNs and the local correlation range, struggle to address these challenges in a global manner \cite{xu2022gmflow}. Transformer-based models with self-attention can indeed address ambiguity problems to some extent by leveraging global feature aggregation. However, they also entail high computational costs, even when working on a 1/8 resolution of the image rather than the full resolution \cite{sui2022craft}.

In this paper, we propose a novel optical flow model, called NeuFlow, for \emph{real-time} optical flow estimation on edge devices while ensuring \emph{high accuracy}.
As shown in Fig.~\ref{optical_flow_vis}, NeuFlow runs at 30fps on a Jetson Orin Nano to process images with the resolution of 512$\times$384.
Specifically, we first use different lightweight CNNs (Convolutional Neural Networks) to encode the input images at different scales of image pyramids. 
They are enhanced by cross-attention to share information between the input images.
Global matching is then adopted at a lower image scale (1/16) to capture large displacement with small computation burden, which is refined by a self-attention module to improve estimation in the ambiguous regions.
The initial optical flow is further processed at the 1/8 scale with CNN layers for local refinement.
It runs much faster than global matching and thus is designed to work on a higher spatial resolution. 
Finally, full-resolution optical flow is obtained through a convex upsampling module.

We conduct experiments on standard benchmarks, training solely on the FlyingChairs and FlyingThings datasets, and evaluate on both the FlyingThings and Sintel datasets for full-resolution flow. Fig. \ref{epe_fps_1_1} shows the end-point error versus frames per second on an RTX 2080. We achieve comparable accuracy to the latest optical flow methods, including RAFT, GMFlow, and GMA, while being 10× faster. FlowFormer achieves the highest accuracy but is 70× slower than our method.

 Our main contribution is an optical flow system. We make design choices that ensure real-time inference on edge devices without postprocessing (\textit{e.g.}, compression, pruning) and high accuracy at the same time.
Our code and model weights have been publicly released.
By sharing NeuFlow with the community, we believe it will empower the next generation of robotic SLAM, visual SLAM and visual inertial odometry applications on UAS and other SWaP-C contained robotic vehicles.

\begin{figure*}[ht]
\begin{center}
\includegraphics[width=\textwidth]{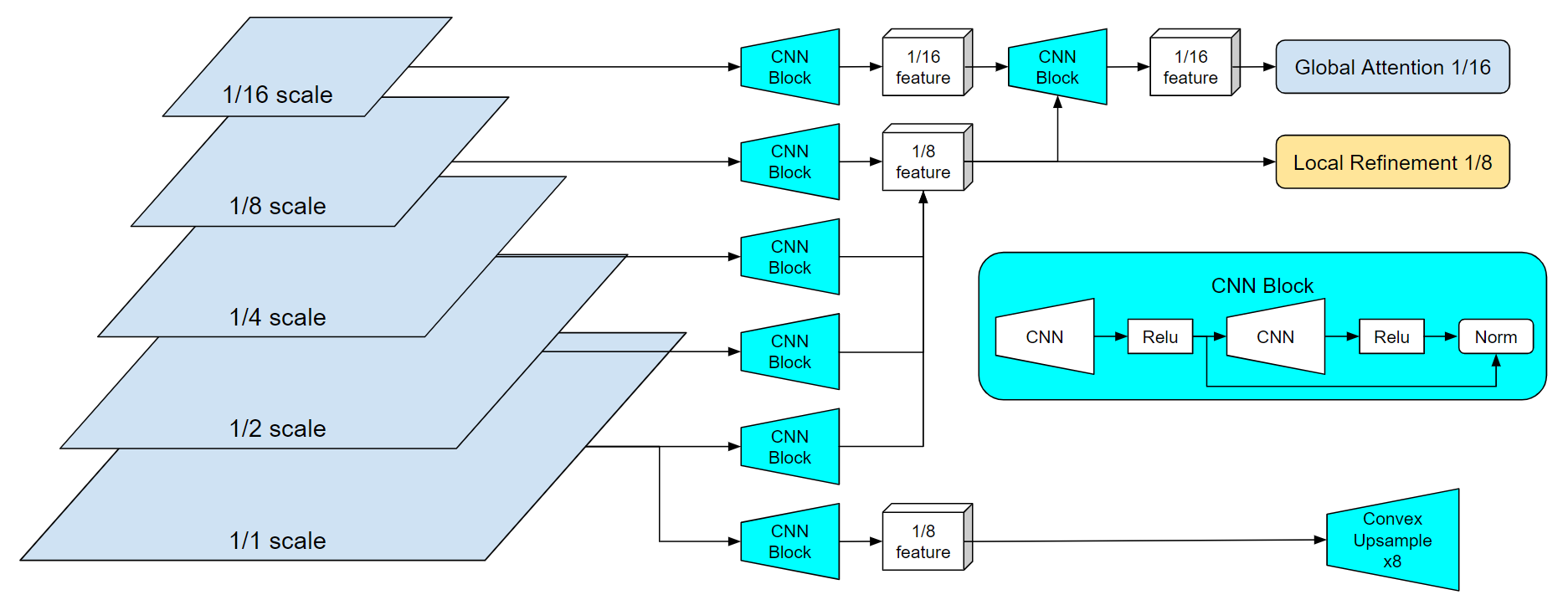}
\end{center}
\caption{NeuFlow Shallow CNN Backbone: Initially, we downsample the image into different scales, ranging from 1/1 scale to 1/16 scale. Subsequently, we extract features using a CNN block. The feature vectors at 1/1, 1/2, 1/4, and 1/8 scales are concatenated into a single 1/8 feature vector. Then, another CNN block is employed to merge the 1/8 feature vector with the 1/16 feature vector, resulting in a 1/16 feature vector. The 1/16 feature vector is utilized for global attention, while the 1/8 feature vector is employed for local refinement. The CNN block consists solely of two CNN layers along with activation functions and normalization layers. The kernel size and stride of the CNN layers depend on the input and output dimensions upstream and downstream of the network. An additional 1/8 feature is extracted from the full-resolution image to perform convex upsampling.
}
\label{neuflow_backbone}
\end{figure*}

\section{Related Work}

FlowNet \cite{dosovitskiy2015flownet} was the first end-to-end convolutional network for optical flow estimation, proposing two variants: FlowNetS and FlowNetC, along with the synthetic FlyingChairs dataset for end-to-end training and benchmarking. An improved version, FlowNet 2.0 \cite{ilg2017flownet}, fused cascaded FlowNets with a small displacement module and decreased the estimation error by more than 50\%.

Following FlowNet \cite{dosovitskiy2015flownet}, researchers sought lightweight optical flow methods. SPyNet \cite{ranjan2017optical} computed optical flow by combining a classical spatial-pyramid formulation, offering a model 96\% smaller than FlowNet in terms of model parameters. PWC-Net \cite{sun2018pwc} was 17 times smaller in size than the FlowNet 2 \cite{ilg2017flownet} model. LiteFlowNet \cite{hui2018liteflownet} presented an alternative network with comparable results to FlowNet 2 \cite{ilg2017flownet} while being 30 times smaller in model size and 1.36 times faster in running speed. LiteFlowNet 2 \cite{hui2020lightweight} improved optical flow accuracy on each dataset by around 20\% while being 2.2 times faster. LiteFlowNet 3 \cite{hui2020liteflownet3} further improved flow accuracy by exploring local flow consistency. VCN \cite{yang2019volumetric} utilizes volumetric encoder-decoder architectures to efficiently capture large receptive fields, reducing computation and parameters while preserving accuracy. DCFlow \cite{xu2017accurate} estimates accurate optical flow through direct cost volume processing. DCVNet \cite{jiang2023dcvnet}, a novel approach that combines dilated cost volumes and 3D convolutions, has demonstrated real-time inference on 1080ti GPU while achieving comparable accuracy to other optical flow estimation approaches.

More recently, RAFT \cite{teed2020raft} used recurrent all-pairs field transforms to achieve strong cross-dataset generalization as well as high efficiency in inference time, training speed, and parameter count. GMA \cite{jiang2021learning} used global motion aggregation to help resolve ambiguities caused by occlusions. GmFlow \cite{xu2022gmflow} reformulated optical flow as a global matching problem to achieve both high accuracy and efficiency. CRAFT \cite{sui2022craft} used a cross-attentional flow transformer to revitalize the correlation volume computation. FlowFormer \cite{huang2022flowformer}, \cite{shi2023flowformer++} introduced a transformer-based neural network architecture for learning optical flow and has achieved new state-of-the-art performance. Many works \cite{xu2021high}, \cite{jiang2021learning2}, \cite{zhang2021separable}, \cite{hofinger2020improving}, \cite{hur2019iterative}, \cite{wang2020displacement}, \cite{truong2020glu} are also proposed to either reduce the computational costs or improve the flow accuracy.

\begin{table*}[ht]
\begin{center}
\begin{tabular}{|c|c|c|c|c|c|c|}
\hline
Full Res (960×540) & Things (val, clean) & Things (val, final) & RTX 2080 (s) & Jetson Orin Nano (s) & Batch Size (8G) & Params \\
\hline
FlowFormer & 3.488 & \textbf{2.774} & 0.834 & N/A & 2 & 16.17M \\
FlowFormer (small) & 9.773 & 8.745 & 0.661 & N/A & 2 & 6.18M \\
GMFlow (1 iter) & \textbf{3.271} & 3.123 & 0.115 & 1.000 & 8 & 4.68M \\
RAFT (12 iters) & 4.122 & 3.775 & 0.142 & 0.878 & 8 & 5.26M \\
GMA (6 iters) & 4.396 & 3.785 & 0.145 & 0.936 & 4 & 5.88M \\
\hline
NeuFlow & 3.846 & 3.828 & \textbf{0.013} & \textbf{0.097} & \textbf{42} & \textbf{3.85M} \\
\hline
Full Res (1024×436) & Sintel (train, clean) & Sintel (train, final) & RTX 2080 (s) & Jetson Orin Nano (s) & Batch Size (8G) & Params \\
\hline
FlowFormer & \textbf{1.004} & \textbf{2.401} & 0.715 & N/A & 2 & 16.17M \\
FlowFormer (small) & 1.324 & 2.679 & 0.534 & N/A & 2 & 6.18M \\
GMFlow (1 iter) & 1.495 & 2.955 & 0.097 & 0.820 & 10 & 4.68M \\
RAFT (12 iters) & 1.548 & 2.791 & 0.124 & 0.760 & 8 & 5.26M \\
GMA (6 iters) & 1.423 & 2.866 & 0.141 & 0.747 & 8 & 5.88M \\
\hline
NeuFlow & 1.660 & 3.126 & \textbf{0.011} & \textbf{0.084} & \textbf{64} & \textbf{3.85M} \\
\hline
\end{tabular}
\end{center}
\caption{This table compares latest optical flow methods when outputting full-resolution flow, all models are trained with FlyingThings and FlyingChairs: FlowFormer achieves the highest accuracy but is 70 times slower than NeuFlow. GmFlow achieves 20\% higher accuracy than NeuFlow on the FlyingThings dataset and demonstrates similar accuracy on Sintel; however, NeuFlow is 10 times faster. Compared to RAFT (12 iters) and GMA (6 iters), NeuFlow achieves comparable accuracy on both the FlyingThings and Sintel datasets, while being 12 times faster. Additionally, the batch size indicates that NeuFlow consumes less memory.}
\label{table_1}
\end{table*}

\begin{table*}[ht]
\begin{center}
\begin{tabular}{|c|c|c|c|c|c|}
\hline
1/8 Res (120×66) & Things (val, clean) & Things (val, final) & RTX 2080 (s) & Jetson Orin Nano (s) & Batch Size (8G) \\
\hline
FlowFormer & 0.463 & \textbf{0.394} & 0.819 & N/A & 2 \\
FlowFormer (small) & 1.244 & 1.111 & 0.647 & N/A & 2 \\
GMFlow (1 iter) & \textbf{0.434} & 0.411 & 0.114 & 0.994 & 8 \\
RAFT (12 iters) & 0.574 & 0.527 & 0.136 & 0.830 & 10 \\
GMA (6 iters) & 0.608 & 0.528 & 0.142 & 0.914 & 6 \\
\hline
NeuFlow & 0.525 & 0.518 & \textbf{0.010} & \textbf{0.078} & \textbf{56} \\
\hline
1/8 Res (128×54) & Sintel (train, clean) & Sintel (train, final) & RTX 2080 (s) & Jetson Orin Nano (s) & Batch Size (8G) \\
\hline
FlowFormer & \textbf{0.145} & \textbf{0.313} & 0.700 & N/A & 2 \\
FlowFormer (small) & 0.195 & 0.355 & 0.548 & N/A & 2 \\
GMFlow (1 iter) & 0.188 & 0.367 & 0.096 & 0.816 & 10 \\
RAFT (12 iters) & 0.217 & 0.365 & 0.118 & 0.747 & 14 \\
GMA (6 iters) & 0.198 & 0.370 & 0.139 & 0.733 & 8 \\
\hline
NeuFlow & 0.220 & 0.394 & \textbf{0.008} & \textbf{0.068} & \textbf{72} \\
\hline
\end{tabular}
\end{center}
\caption{This table compares latest optical flow methods when outputting 1/8 resolution flow, all models are trained with FlyingThings and FlyingChairs: NeuFlow is optimized for higher accuracy and efficiency at 1/8 resolution flow. We achieve significantly higher accuracy than RAFT (12 iters) and GMA (6 iters) on the FlyingThings dataset. Additionally, NeuFlow is 80 times faster than FlowFormer and 12 times faster than GmFlow, RAFT, and GMA on both GPU platforms.}
\label{table_2}
\end{table*}

\section{Proposed Approach: NeuFlow}

We introduce NeuFlow, a global-to-local architecture for optical flow that achieves both high accuracy and efficiency. 
% This architecture implements a global-to-local approach to efficiently tackle optical flow challenges. 
An illustration of NeuFlow's architecture is shown in Fig. \ref{neuflow_arct}.
Initially, a shallow CNN backbone extracts low-level features from a multi-scale image pyramid. Next, global cross-attention and self-attention are applied at the 1/16 scale to address the challenges of large displacement. Subsequently, local refinement is conducted at the 1/8 scale to yield high-accuracy optical flow. Convex upsampling is then employed to generate full-resolution flow. 

\subsection{Shallow CNN Backbone}
While most optical flow methods employ a relatively deep CNN backbone for feature extraction, we believe that high-level, semantical encoding of input images is not necessary for optical flow tasks. Instead, sufficient low-level features are more crucial. Thus, in our approach, we employ a simple CNN block to directly extract features at multiple scales of the images, as depicted in Fig. \ref{neuflow_backbone}.

Simple CNN blocks are used to extract features from various scales of the input images.
Each block comprises only two CNN layers, activation functions, and normalization. This design prioritizes the extraction of a large number of low-level features directly from the image. High-level CNN is only employed to merge features from different scales.

\subsection{Global Cross-Attention}
Similar to GMFlow~\cite{xu2022gmflow}, we utilize Transformers~\cite{vaswani2017attention} to implement global cross-attention. This mechanism takes features from image one as the query and features from image two as both key and value. This mechanism enhances the distinctiveness of matching features and reduces the similarity of unmatched features.

Global matching is then applied to find corresponding features. Unlike local regression-based optical flow methods, this approach does not restrict the range of flow between image pairs. Consequently, it performs well on datasets with large pixel displacement, such as FlyingThings \cite{mayer2016large}, and exhibits greater stability in real-world scenarios, including fast-moving camera situations.

However, global attention tends to be significantly slow as it computes correlations between all pixels in the image. Due to this heavy computational load, many transformer-based optical flow methods operate at a lower resolution (1/8), but it remains too slow. In our approach, we implement global cross-attention on 1/16 resolution features and stack 2 layers of it. Additionally, we apply Flash-attention \cite{dao2022flashattention} for slight speed improvement.

\subsection{Flow Self-Attention}
The cross-attention mechanism for pixel matching operates under the assumption that all matching pixels are visible in the image pair. However, this assumption is not always accurate. Out-of-boundary pixels and occluded pixels can violate this assumption, posing a significant ambiguity challenge in optical flow estimation. To address this issue, we incorporate global self-attention on features to globally assess the similarity of pixels. This allows us to propagate unseen flows based on such similarity. Additionally, the implementation of this process can be optimized using flash-attention for improved speed.

\subsection{Local Refinement}
As the cross-attention at the 1/16 scale of the image has already established the global correspondence of pixels, we focus on local refinement at a larger scale, specifically 1/8 in our case. Initially, we warp the features of image two using the flow computed at 1/16 scale, ensuring that matching pixels in the image pair are located nearby within a small range. To determine the best matching pixel within this range, we compute the local correlation of each pixel in image one with the nearby 7x7 pixels on the warped image two. The feature vector and the estimated coarse flow are also incorporated and fed into deep CNN layers to estimate the delta flow at the 1/8 scale.

\subsection{Upsampling Module}
Similar to the latest optical flow methods like GmFlow and RAFT, we adopt a scheme that estimates optical flow at 1/8 resolution and then upsamples the flow to full resolution. The upsampling module resembles RAFT's approach, where each pixel of the high-resolution flow field is determined as the convex combination of its 9 coarse resolution neighbors using weights predicted by the network. However, instead of utilizing features employed for matching at 1/16 scale and 1/8 scale, we directly extract features from the original image using a simple CNN block, as illustrated in Fig. \ref{neuflow_backbone}. This approach allows us to obtain feature maps with finer details, thereby sightly enhancing the accuracy of the full-resolution flow, albeit at the expense of additional computation time.

\begin{table*}[ht]
\begin{center}
\begin{tabular}{|c|c|c|c|c|c|}
\hline
Full Res (960×540) & Things (val, clean) & Things (val, final) & RTX 2080 (s) \\
\hline
FlowNet 2 (pytorch) & 6.782 & 6.774 & 0.091 \\
PWC-Net (pytorch) & 8.098 & 8.168 & 0.033 \\
LiteFlowNet (pytorch) & 9.033 & 9.018 & 0.072 \\
\hline
NeuFlow & \textbf{3.846} & \textbf{3.828} & \textbf{0.013} \\
\hline
Full Res (1024×436) & Sintel (train, clean) & Sintel (train, final) & RTX 2080 (s) \\
\hline
FlowNet 2 (pytorch) & 2.222 & 3.639 & 0.085 \\
PWC-Net (pytorch) & 2.643 & 4.060 & 0.029 \\
LiteFlowNet (pytorch) & 2.588 & 4.058 & 0.059 \\
\hline
NeuFlow & \textbf{1.660} & \textbf{3.126} & \textbf{0.011} \\
\hline
\end{tabular}
\end{center}
\caption{This table compares NeuFlow with local regression-based optical flow methods, all models are trained with FlyingThings and FlyingChairs: NeuFlow consistently demonstrates a significant advantage in both accuracy and efficiency on both the FlyingThings and Sintel datasets. Inference time is measured using PyTorch implementations of each model.}
\label{table_3}
\end{table*}

\begin{table*}[ht]
\begin{center}
\begin{tabular}{|c|c|c|c|c|c|}
\hline
Full Res (960×540) & Things (val, clean) & Things (val, final) & RTX 2080 (s) \\
\hline
LiteFlowNet 2 & 10.395 & 10.205 & N/A \\
LiteFlowNet 3 (pytorch) & 9.856 & 9.692 & 0.050 \\
\hline
NeuFlow & \textbf{4.044} & \textbf{4.025} & \textbf{0.013} \\
\hline
Full Res (1024×436) & Sintel (train, clean) & Sintel (val, final) & RTX 2080 (s) \\
\hline
LiteFlowNet 2 & 1.559 & 1.944 & N/A \\
LiteFlowNet 3 (pytorch) & 1.451 & 1.920 & 0.042 \\
\hline
NeuFlow & \textbf{0.987} & \textbf{1.294} & \textbf{0.011} \\
\hline
\end{tabular}
\end{center}
\caption{This table compares NeuFlow with LiteFlowNet 2 and 3. As these models do not provide models trained solely on the C+T dataset, we compare them with models trained with mixed datasets. NeuFlow consistently demonstrates a significant advantage in both accuracy and efficiency on both the FlyingThings and Sintel datasets.}
\label{table_4}
\end{table*}

\section{Experiments}
\subsection{Training and Evaluation Datasets}
The common optical flow training process is typically divided into three stages: Stage one involves training on the FlyingChairs dataset, followed by stage two, which entails training on the FlyingThings dataset. Stage three involves training on a mixed dataset comprising Sintel \cite{butler2012naturalistic}, Kitti \cite{menze2015object}, and HD1K. We follow the same procedure, utilizing the training and evaluation code derived from FlowNet 2.

To ensure a fair comparison of different optical flow models, we compare the training results of stage two (FlyingThings) rather than stage three (Sintel+Kitti+HD1K). This decision is motivated by several factors. Firstly, different researchers may use varying training ratios for different datasets, potentially biasing results towards models trained more heavily on specific datasets. Additionally, the limited submission of these datasets complicates the validation process.

Since stage two only involves training on FlyingChairs and FlyingThings train set, we validate the model on both the FlyingThings test set and the Sintel training set. The FlyingThings dataset presents challenges due to its large displacement, which tests the model's ability to estimate fast-moving optical flow. Meanwhile, validation on the Sintel dataset helps demonstrate the model's cross-dataset generalization, as it has not been trained on data from the same domain.

\subsection{Comparision with Latest Optical Flow Methods}
We begin by comparing our method to several state-of-the-art optical flow methods renowned for their superior accuracy: RAFT, GMA, GmFlow, and FlowFormer (Table. \ref{table_1}). Unfortunately, due to its computational demands, CRAFT could not be evaluated on our RTX 2080. Our comparison focuses on accuracy differences across the Sintel and FlyingThings datasets, as well as computation time on both the RTX 2080 and the edge computing platform Jetson Orin Nano, considering image sizes of 1024×436 for Sintel and 960×540 for FlyingThings. The inference batch size is also measured on an 8GB GPU memory to assess the memory usage of different models.

Among these methods, FlowFormer achieves the highest accuracy. However, it is significantly slower than our approach, being approximately 70 times slower on the RTX 2080. Although FlowFormer offers a smaller model version, it fails to perform adequately on the large displacement dataset FlyingThings and remains approximately 60 times slower than our method. Unfortunately, FlowFormer couldn't be implemented on Jetson Orin Nano for evaluation.

GmFlow exhibits slightly better accuracy than our method on datasets with large displacement (FlyingThings) due to its global attention mechanism operating on a 1/8 image scale, whereas ours operates on a 1/16 scale. Specifically, GmFlow shows a 20\% accuracy improvement over our method on the FlyingThings dataset and comparable performance on the Sintel dataset. However, our method excels in efficiency, being roughly 10 times faster than GmFlow on both GPU platforms across various resolutions.

Both RAFT and GMA utilize multiple iterations to refine optical flow estimates. We consider 12 iterations for RAFT and 6 iterations for GMA, as they achieve similar accuracy compared to our method at such iterations. Notably, on the FlyingThings dataset, characterized by large displacement, our method outperforms both RAFT and GMA on clean sets and performs comparably on final sets. On the Sintel dataset, where pixel movement is relatively smaller, our method achieves similar accuracy. These results underscore the effectiveness of our global attention mechanism in handling large displacement scenarios, while also exhibiting a 12 times speedup over RAFT and GMA on both GPUs at various resolutions. To ensure fair comparison of computation time, we disable mixed precision computation.

\subsection{Comparison on 1/8-resolution Flow}
In certain applications such as SLAM, it's common practice to use 1/8 image resolution instead of full resolution to reduce computational burden for downstream operations. Recent optical flow methods adopt the same approach to obtain 1/8 resolution flow and employ an upsampling module to provide full resolution flow. Our approach follows this scheme and is specifically optimized for 1/8 image resolution to meet real-time requirements on edge computing platforms.

Table. \ref{table_2} illustrates the comparison results. Compared to full resolution, our approach achieves relatively higher accuracy with less computation time. For example, on the FlyingThings dataset, we significantly outperform RAFT (12 iters) and GMA (6 iters) on 1/8 resolution flow, whereas our performance advantage is not as pronounced on full resolution flow.

\subsection{Comparision with Local Regression-based Optical Flow Methods}
We then compared our approach with local regression-based CNN optical flow methods, encompassing popular methods such as FlowNet 2, PWC-Net, and the LiteFlownet series (LiteFlownet 1, LiteFlownet 2, LiteFlownet 3) (Table. \ref{table_3}). Due to the limited receptive field of CNNs and the constrained range of local correlation, none of these methods performed adequately on the large displacement dataset FlyingThings, despite being trained on it. Our approach also consistently outperforms these local regression optical flow methods on Sintel dataset, demonstrating an approximate 40\% accuracy improvement.

Local regression-based optical flow methods generally offer faster speeds compared to the latest high-accuracy optical flow methods. To ensure a fair comparison of computation time with our approach, we opted to use their PyTorch implementations instead of Caffe. The results reveal that PWC-Net emerges as the fastest method among these local regression approaches. However, it also exhibits the lowest accuracy and remains approximately 3 times slower than our approach on the RTX 2080. FlowNet 2, despite its better accuracy, operates around 8 times slower than our approach.

The LiteFlowNet series also lags behind in speed, with a factor of 4 to 6 slower than ours. Only LiteFlowNet 1 provides a model trained specifically with the FlyingThings dataset (Stage 2), whereas LiteFlowNet 2 and 3 offer models trained with mixed datasets (Stage 3). Additionally, Table. \ref{table_4} illustrates that LiteFlowNet 2 and 3 fail to perform adequately on the large displacement dataset (FlyingThings), even though they were trained on it. We trained our model on mixed datasets (Stage 3) for a fair comparison, and we achieved a significant advantage in both accuracy and computation time on the RTX 2080. While we couldn't find the PyTorch implementation for LiteFlowNet 2, published data suggests that its speed is comparable to that of PWC-Net. Unfortunately, we failed to build these methods on Jetson Orin Nano due to its support for only the latest PyTorch version.

\begin{figure}[t]
\begin{center}
\includegraphics[width=\linewidth]{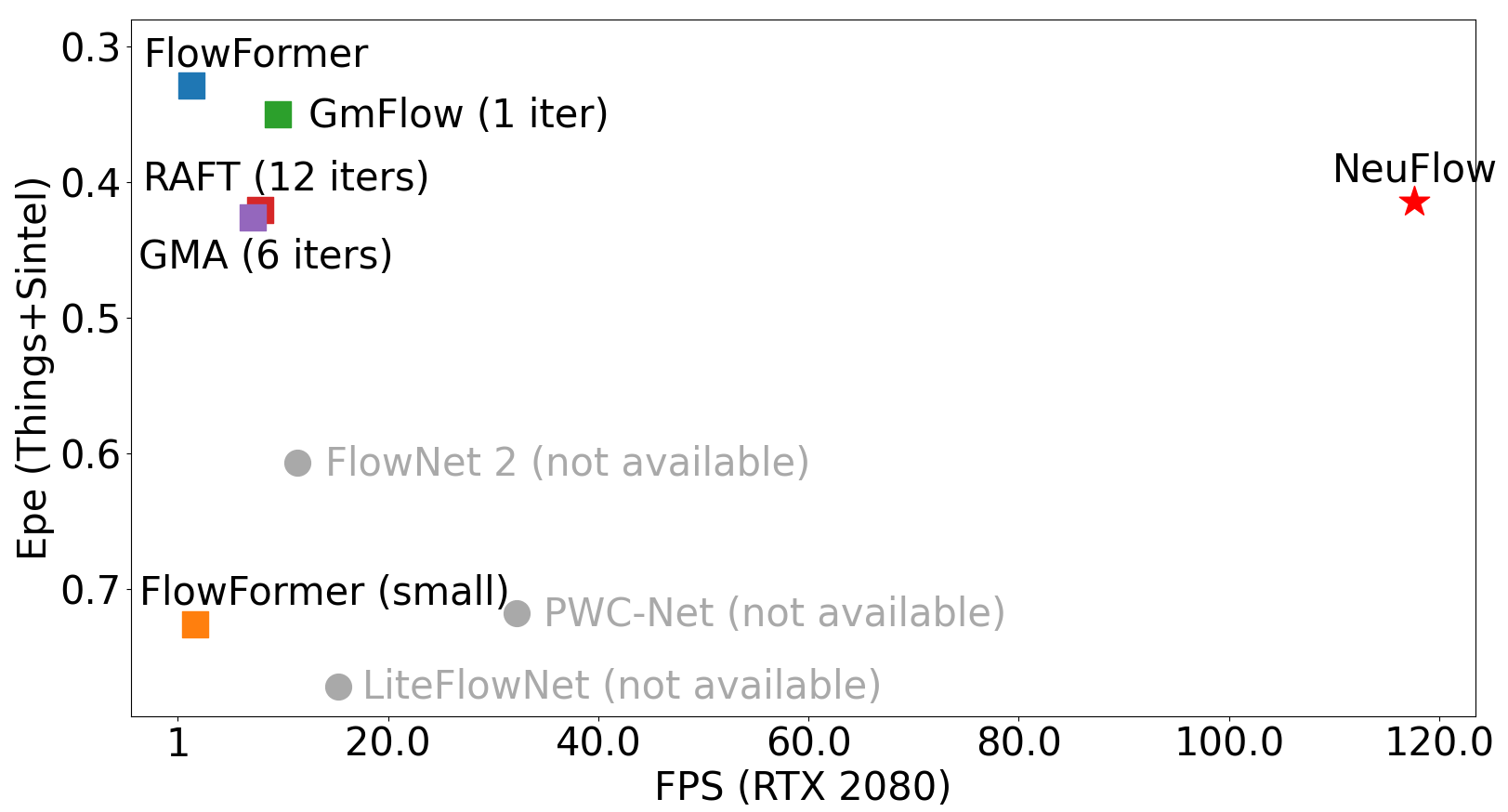}
\end{center}
\caption{End point error (EPE) v.s. frame per second (FPS) on Nvidia RTX 2080 while outputting 1/8 resolution flow. All models trained solely on FlyingThings and FlyingChairs. NeuFlow is optimized for accuracy and efficiency at 1/8 resolution, thus we gain more advantage compared to full resolution flow.}
\label{epe_fps_1_8}
\end{figure}

\subsection{Overall Comparison}

We plot the end-point error (EPE) versus frames per second (FPS) throughput on Nvidia RTX 2080 (see Fig. \ref{epe_fps_1_1}). Each point represents an optical flow method. All models are trained with FlyingChairs and FlyingThings datasets. EPE is measured by averaging EPE values of Things Clean, Things Final, Sintel Clean, and Sintel Final equally. Inference time is measured by averaging the running time on Things images (960×540) and Sintel images (1024×436). We observe that NeuFlow achieves comparable accuracy to the latest optical flow methods while being 10×-70× faster. Compared to local regression-based methods, we have significant advantages in both accuracy and efficiency. Since we optimize for 1/8 resolution flow, we gain more speed advantage and accuracy compared to full resolution (see Fig. \ref{epe_fps_1_8}) Since local regression-based optical flow methods do not adhere to the same scheme of high accuracy 1/8 flow followed by an 8x upsampling module, resulting in lower accuracy for 1/8 resolution flow, we have omitted them from this plot.

\subsection{Inference Time on Jetson Orin Nano}

As our ultimate goal is to develop an optical flow method capable of real-time performance on edge computing platforms, we measured the inference frames per second (FPS) on Jetson Orin Nano at various image resolutions (Table. \ref{table_5}). For applications utilizing 1/8 resolution optical flow, such as SLAM, we also measured the FPS of it. Since most vision applications process image streams and estimate optical flow on continuous frames, the backbone of the previous frame has already been computed, which can be utilized for the optical flow estimation of the next continuous two frames. Therefore, we also measured the FPS when feeding only one frame into the backbone neural network. The results show that we are able to achieve around 30 FPS on smaller images when outputting full resolution optical flow, and 36 FPS on larger images when outputting 1/8 resolution flow.

\begin{table}[t]
\begin{center}
\begin{tabular}{|c|c|c|c|}
\hline
 & \makecell{Things 1× \\(960×540)} & \makecell{Sintel 1× \\(1024×436)} & \makecell{Chairs 1× \\(512x384)} \\
\hline
Inference on 2 frames & 10.3 & 11.9 & 25.0 \\
Inference on 1 frame & 11.9 & 13.9 & 29.9 \\
\hline
 & \makecell{Things 1/8× \\(120×66)} & \makecell{Sintel 1/8× \\(128×54)} & \makecell{Chairs 1/8× \\(64×48)} \\
\hline
Inference on 2 frames & 12.8 & 14.7 & 29.4 \\
Inference on 1 frame & 15.4 & 17.9 & 36.4 \\
\hline
\end{tabular}
\end{center}
\caption{NeuFlow achieves real-time performance on Jetson Orin Nano for specific image resolutions, with 1/8 resolution flow offering faster inference times. In image stream processing, only one frame needs backbone computation, as the other frame is already computed in the preceding one.}
\label{table_5}
\end{table}

\section{Conclusions and Future Work}

In this paper, we proposed a novel optical flow architecture called NeuFlow, which enables real-time optical flow estimation on edge computing platforms like Jetson Orin Nano. NeuFlow is 10×-80× faster than the latest optical flow methods, with comparable accuracy on both the FlyingThings and Sintel datasets. Therefore, NeuFlow ensures better performance across various use cases. We have released the code and model weights of NeuFlow (https://github.com/neufieldrobotics/NeuFlow) to allow the community full access to use, modify and experiment with as they see fit. 

However, we also recognize that sacrificing some computation time for higher accuracy may be necessary for certain users. Conversely, further improvement in efficiency is also possible. Thus, many options are there to extend the model and achieve higher accuracy or higher efficiency, which we leave as future work.

\textbf{Higher Accuracy.}
Extending the model in various ways can be effective, such as expanding the feature dimension (ours is 90), stacking more cross-attention layers, increasing the depth of the CNN backbone, or adding CNN layers in the local refinement step. Additionally, iterative refinement is also an option to improve accuracy.

Local refinement can also be applied at 1/16 resolution to refine flow at this lower resolution, which can propagate improved flow to higher resolutions. Moreover, global attention and local refinement can be utilized at higher resolutions. For instance, similar to GmFlow, one can perform global cross-attention at 1/8 resolution and refine at 1/4 resolution, promising improved accuracy across all resolution flows.

\textbf{Higher Efficiency.} Our approach consistently employs native CNN architectures, yet there are several more efficient CNN architectures available that can further enhance efficiency. For instance, MobileNets \cite{howard2017mobilenets}, \cite{sandler2018mobilenetv2}, \cite{howard2019searching} leverage depthwise separable convolutions to construct lightweight deep neural networks, while ShuffleNet \cite{zhang2018shufflenet} utilizes pointwise group convolution and channel shuffle techniques to reduce computation costs while maintaining accuracy.

Other techniques, such as NVIDIA TensorRT, offer low latency and high throughput for optimized runtime performance. Mixed precision techniques, using 16-bit or lower precisions during training and inference, can significantly speed up inference and reduce memory usage on modern GPUs. Network pruning is also effective in reducing the size of heavy networks by removing redundant parameters while maintaining comparable accuracy.

\bibliographystyle{IEEEtran}
\bibliography{IEEEabrv,mybib}

\end{document}